\documentclass{llncs}

\usepackage[a4paper,includeheadfoot,top=1.65in,bottom=1.65in,left=1.73in,hcentering]{geometry}
\usepackage[pdftex]{graphicx}
\usepackage[sectionbib]{natbib}
\usepackage{caption} 
\usepackage[utf8]{inputenc}
\usepackage{t1enc}
\usepackage{tikz-dependency}
\usepackage{qtree}
\usepackage{float}
\usepackage[title]{appendix}

\usepackage{booktabs}
\usepackage{array}
\usepackage{multirow}
\usepackage{amsmath}
\usepackage{commath}
\usepackage{tabularx,booktabs}

\usepackage{subcaption}
\restylefloat{figure}
\usepackage{tikz-qtree}
\usepackage[english]{babel}
\renewcommand\bibname{References}

\newif{\ifhidecomments}
\hidecommentsfalse

\ifhidecomments
    \newcommand{\bence}[1]{}
    \newcommand{\attila}[1]{}
    \newcommand{\judit}[1]{}
    \newcommand{\patrick}[1]{}
    \newcommand{\balazs}[1]{}
\else
    \newcommand{\bence}[1]{\textcolor{red}{[#1 ({\bf Bence})]}}
    \newcommand{\attila}[1]{\textcolor{blue}{[#1 ({\bf Attila})]}} 
    \newcommand{\judit}[1]{\textcolor{orange}{[#1 ({\bf Judit})]}} 
    \newcommand{\patrick}[1]{\textcolor{green}{[#1 ({\bf Patrick})]}} 
    \newcommand{\balazs}[1]{\textcolor{pink}{[#1 ({\bf Balazs})]}} 

\fi

\begin{document}

\title{Syntax-based data augmentation for Hungarian-English machine translation}
	\author{Attila Nagy\inst{1}, Patrick Nanys\inst{1}, Bal\'azs Frey Konr\'ad\inst{1}, Bence Bial\inst{1}, Judit \'Acs\inst{2}\\
\institute{
$^1$Department of Automation and Applied Informatics\\
Budapest University of Technology and Economics\\
$^2$Institute for Computer Science and Control\\
Eötvös Loránd Research Network}
}

\maketitle

\begin{abstract}
We train Transformer-based neural machine translation models for Hungarian-English and English-Hungarian using the Hunglish2 corpus. Our best models achieve a BLEU score of 40.0 on Hungarian-English and 33.4 on English-Hungarian. Furthermore, we present results on an ongoing work about syntax-based augmentation for neural machine translation. Both our code and models are publicly available\footnote{https://github.com/attilanagy234/syntax-augmentation-nmt}.
\end{abstract}

\section{Introduction}
Machine Translation (MT) is a subfield of natural language processing, where the task is to perform translation automatically from one language to another. To be able to translate corpora between arbitrary languages, one needs to develop a deep understanding of the underlying structure of language and for this reason MT has been considered one of the hardest problems in NLP.

Our contributions with this work are twofold. First, we train neural machine translation models for English-Hungarian and Hungarian-English using Transformer models. Second, we propose a new data augmentation technique for machine translation using syntactic parsing. To the best of our knowledge, no prior work has been published on Transformer-based machine translation for Hungarian-English.

\section{Related work}
Early approaches tried to model translation by deriving translation rules based on our knowledge of linguistics. A rule-based method, however, is insufficient for covering the countless edge cases in language. With the increasingly available parallel datasets, data-driven approaches gained dominance in the previous decades. Statistical machine translation (SMT) \citep{koehn2003statistical, brown1990statistical} outperforms rule-based methods by learning latent structures in the data with the help of statistical methods. Although better than its predecessor, SMT struggles to capture long-term dependencies. Neural machine translation (NMT) \citep{bahdanau2014neural, cho2014learning} tackles this problem by modeling translation as an end to end process using neural networks. In current machine translation research, the Transformer architecture \citep{vaswani2017attention} is almost exclusively used in supervised settings \citep{tran2021facebook, germann2021university, oravecz2020etranslation}. For the Hungarian-English language-pair, published methods followed the same evolution of rule-based systems \citep{proszeky2002metamorpho}, statistical methods \citep{laki2013english} and neural models \citep{tihanyi2017first}.

Data augmentation is particularly important in machine translation research, because many language-pairs have insufficient resources to build complex models. Classical augmentation methods that are used in NLP are hard to apply to machine translation, because it is very hard to augment both the source and target sentence such that the parallelism of the sentence-pair holds. \cite{wang2015s} stochastically select words for replacement based on a a distance metric in an embedding space. \cite{kobayashi2018contextual} train a language model to predict new words based on its surrounding context and use this model to replace words. \cite{xie2017data} avoid overfitting to specific contexts by randomly replacing words in the training data with a blank token. Methods that use the dependency parse tree of a sentence for augmentation were proved useful in a number of tasks such as word relation classification \citep{xu2016improved}, POS tagging \citep{csahin2019data} and dependency parsing \citep{vania2019systematic, dehouck2020data}. \cite{duan2020syntax} use the depth of words in a dependency parse tree as a clue of importance for selecting words for augmentation in machine translation.
The syntax-aware data augmentation that we discuss in this work was first proposed by \cite{nagy2021developing}. In machine translation, backtranslation is the most common data augmentation method, which creates pseudo-parallel sentences from monolingual data using a baseline translation model \citep{sennrich2015improving}.

\section{Methodology}
We discuss two main experiments in this work. Firstly, we train competent neural machine translation models based on state-of-the-art architectures for HU-EN and EN-HU and provide a solid baseline for future NMT research in Hungarian by releasing the trained model. Secondly, we propose a novel data augmentation technique for machine translation using dependency parsing. As data augmentation is particularly useful when training data is insufficient, we perform these experiments in a simulated low-resource setting, using a subset of the Hunglish2 corpus. 

\subsection{Formulation}
We formulate machine translation on the sentence level. Given a dataset $\mathcal{D}$ that contains parallel sentences from the source and target language $\boldsymbol{x}, \boldsymbol{y} \in \mathcal{D}$, we define the goal of neural machine translation as estimating the unknown conditional probability $P(\boldsymbol{y}|\boldsymbol{x})$. This is a classical sequence-to-sequence problem: an encoder can be used to create a representation of the source sentence, which is fed into a decoder with the purpose of extracting relevant information from this representation. The decoder then generates the output symbols from left to right. This way the decoder can be thought of as a language model conditioned on the output of the encoder and the already generated symbols.

\subsection{Dataset}
We use the Hunglish2 corpus for our experiments, which is a sentence-aligned corpus consisting of 4.1M Hungarian-English bisentences \citep{varga2007parallel}. The dataset was constructed by scraping and aligning bilingual data in several domains from the internet. The distribution of each domain that make up Hunglish2 is shown in Table \ref{table:hunglish2-stats}.

\begin{table}[!htbp]
  \centering
    \begin{tabular}{ l c c }
    \hline
    \textbf{Subcorpus} & \textbf{Tokens} & \textbf{Bisentences} \\
    \hline
    Modern literature & 37.1M & 1.67M \\
    Classical literature & 17.2M & 652k \\
    Movie subtitles & 3.2M & 343k \\
    Software docs & 1.2M & 135k \\
    Legal text & 56.6M & 1.351M \\
    \hline
    Total & 115.3M & 4.151M \\
    \hline
    \end{tabular}
  \caption{Statistics of the Hunglish2 corpus}
  \label{table:hunglish2-stats}
\end{table}

\begin{figure}[!htbp]
    \centering
    \includegraphics[width=0.95\textwidth]{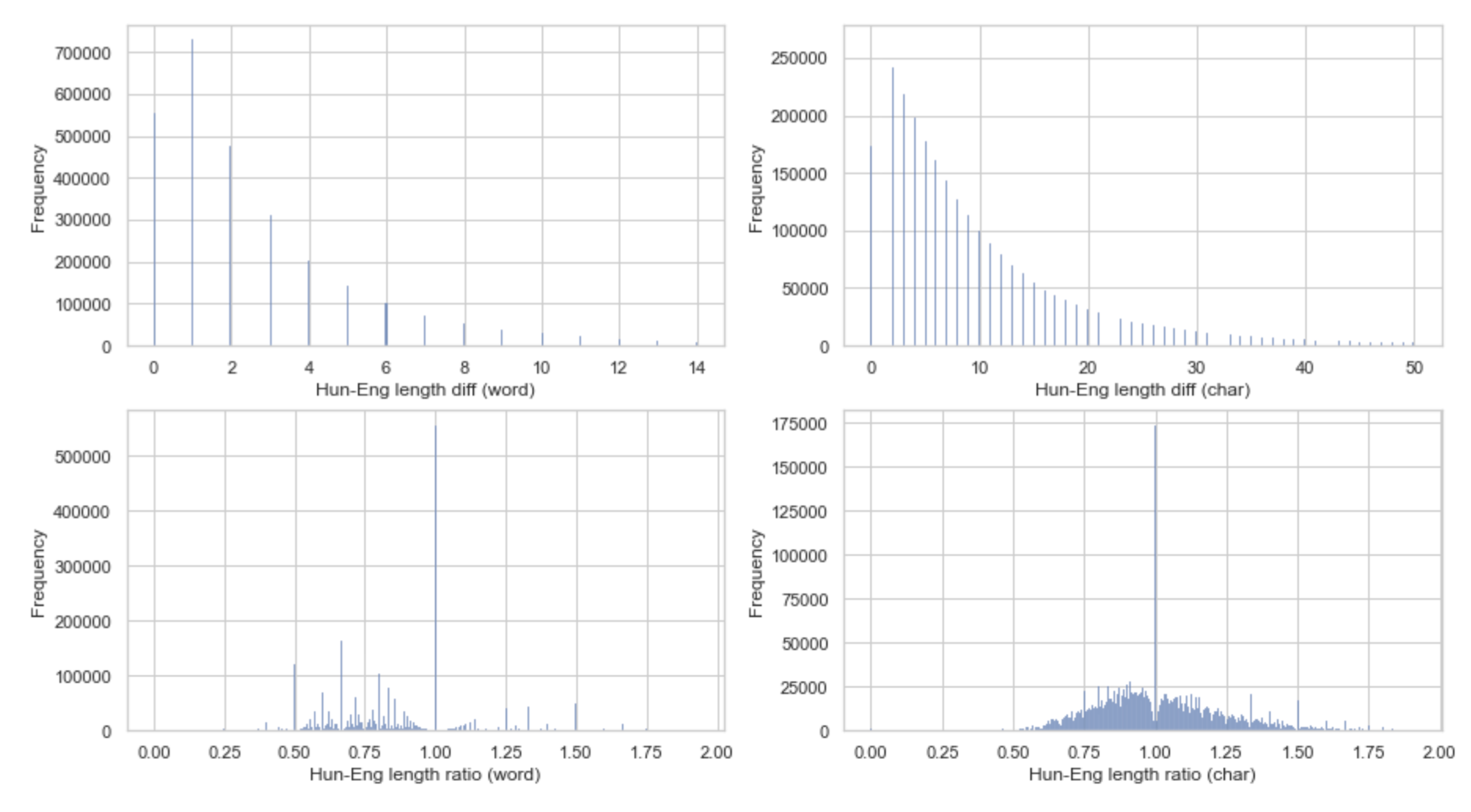}
    \caption{Word- and character-level distributions of the length difference and ratio between source and target sentences in the Hunglish2 corpus.}
    \label{fig:hunglish2-len-ratio-distribution}
\end{figure}

We applied thorough preprocessing before training the models. First, we removed seemingly incorrect data points: sentence pairs, which contained HTML code or were outliers with respect to sentence length. We also remove sentence-pairs, where either side is an empty string. A large number of sentences were wrapped in quotation marks, so in order to avoid overfitting on this behaviour, we also remove leading and trailing quotation marks. Next, we filter the dataset with length-based heuristics. We drop sentences if either the source or target sentence contains more than 32 words. Furthermore, we filter with the relative word counts of the source and target sentences using the below rule:
\begin{align}
(\lvert\text{WC(x) - WC(y)} \rvert < 7) \lor (WC(x)/WC(y) < 1.6) \nonumber
\end{align}
where $WC(x)$ and $WC(y)$ are the word counts of the source and target sentences respectively. The threshold parameters were determined by exploratory data analysis and a series of experiments. See Figure \ref{fig:hunglish2-len-ratio-distribution} for the distribution of length difference and length ratio of bisentences in the raw Hunglish2 corpus. The post-processed dataset contains 3.4M bisentences. Finally, we split the data to train, development and test sets, with a 99-0.5-0.5 ratio. We do this with stratified sampling with respect to the the subcorpora in Hunglish2, to ensure that all splits have a similar distribution.

\subsection{Training}
All of our models use identical architecture and hyperparameters: a vanilla Transformer-based encoder-decoder model. We tokenize the input with the unigram sentencepiece subword tokenizer \citep{kudo2018sentencepiece}, which works particularly well with morphologically rich languages, such as Hungarian. We found that using a shared vocabulary of size 32000 yields the best results. We perform gradient descent using the Adam optimizer \citep{Kingma:2014} with an initial learning rate of 2, 8000 warmup steps and Noam learning rate decay. To avoid overfitting, we perform early stopping based on the perplexity computed on the validation set. Due to computational constraints, we only perform a manual search in the hyperparameter space. The complete set of hyperparameters of our best model can be found in Table \ref{table:base-parameters}. All models were trained using the OpenNMT framework \citep{klein-etal-2017-opennmt} with standalone Nvidia Tesla V100 GPUs.


\subsection{Syntax-aware data augmentation}
One of the greatest challenges in data augmentation for machine translation is preserving parallelism. Taking the example from \citet{duan2020syntax} (see Table \ref{table:replacement-aug-error}.) it is visible, that replacing words only on the source side of the bisentence can easily lead to noisy translation pairs. The success of backtranslation can be partially contributed to the fact that it is good at generating parallel bisentences. Backtranslation, however, might not always be an option, especially in low-resource scenarios, where there is not enough parallel data to train a model that can be used for backtranslation in the first place.

\begin{table}[!htbp]
\centering
\begin{tabular}{l|l||l|l}
\textbf{Parameter}          & \textbf{Value}    & \textbf{Parameter}    & \textbf{Value}       \\ \hline
batch type & tokens & batch size & 4096\\
accumulation count & 4 & average decay & 0.0005 \\
train steps & 150000 & valid steps & 5000 \\
early stopping & 4 & early stopping criteria & ppl \\
optimizer & adam & learning rate & 2 \\
warmup steps & 8000 & decay method & noam \\
adam beta2 & 0.998 & max grad norm & 2 \\
label smoothing & 0.1 & param init & 0 \\
param init glorot & true & normalization & tokens \\
max generator batches & 32 & encoder layers & 8 \\
decoder layers & 8 & heads & 16 \\
RNN size & 1024 & word vector size & 1024 \\
Transformer FF & 2096 & dropout steps & 0\\
dropout & 0.1 & attention dropout & 0.1\\
share embeddings & true & position encoding & true
\end{tabular}
\caption{Hyperparameters of our best model}
\label{table:base-parameters}
\end{table}

\begin{figure}[!htbp]
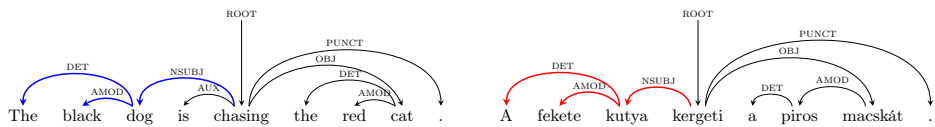

\begin{subfigure}{0.47\textwidth}
\scalebox{.70}{
    \begin{dependency}[theme = simple]
    \centering
      \begin{deptext}[column sep=1em]
          The \& black \& dog \& is \& chasing \& the \& red \& cat \& . \\
      \end{deptext}
      \deproot{5}{ROOT}
      \depedge[arc angle=90, edge style = {blue, thick}]{5}{3}{NSUBJ}
      \depedge[arc angle=50]{5}{4}{AUX}
      \depedge[arc angle=65]{5}{8}{OBJ}
      \depedge[arc angle=90]{5}{9}{PUNCT}
      \depedge[arc angle=90, edge style = {blue, thick}]{3}{1}{DET}
      \depedge[arc angle=30, edge style = {blue, thick}]{3}{2}{AMOD}
      \depedge{8}{6}{DET}
      \depedge[arc angle=30]{8}{7}{AMOD}
    \end{dependency}
}
\end{subfigure}
\hspace*{\fill}
\begin{subfigure}{0.47\textwidth}
\scalebox{.70}{
    \begin{dependency}[theme = simple]
    \centering
      \begin{deptext}[column sep=1em]
          A \& fekete \& kutya \& kergeti \& a \& piros \& macskát \& . \\
      \end{deptext}
      \deproot{4}{ROOT}
      \depedge[edge style = {red, thick}]{4}{3}{NSUBJ}
      \depedge{4}{7}{OBJ}
      \depedge[arc angle=90]{4}{8}{PUNCT}
      \depedge{7}{6}{AMOD}
      \depedge{6}{5}{DET}
      \depedge[arc angle=50, edge style = {red, thick}]{3}{2}{AMOD}
      \depedge[arc angle=90, edge style = {red, thick}]{3}{1}{DET}
    \end{dependency}
}
\end{subfigure}
\caption{An example of finding the same substructure in the dependency trees of a bisentence.}
\label{fig:dep-tree-example}
\end{figure}

\begin{table}[!htbp]
\centering
\begin{tabular}{l|l}
\hline
Original-EN                       & We shall fight on the beaches.    \\ \hline
Original-HU                       & A tengerparton kellene küzdenünk. \\ \hline
Replacement-EN                    & We shall fight with the sandy.    \\ \hline
Replacement-HU (Google Translate) & Harcolni fogunk a homokkal.       \\ \hline
\end{tabular}
\caption{Error analysis of a data augmentation method for NMT \citep{duan2020syntax}}.
\label{table:replacement-aug-error}
\end{table}

We propose a novel syntax-aware data augmentation technique, which is based on a hypothesis, that dependency parse trees of the source and target sentences contain subtrees, which carry the same meaning. If we can identify these subtrees simultaneously in the source and target language (see Figure \ref{fig:dep-tree-example} for an example), we have the possibility to generate new bisentences by swapping subtrees.

\begin{figure}[!htbp]
\centerline{
\scalebox{1.3}{
    \includegraphics[width=0.99\textwidth]{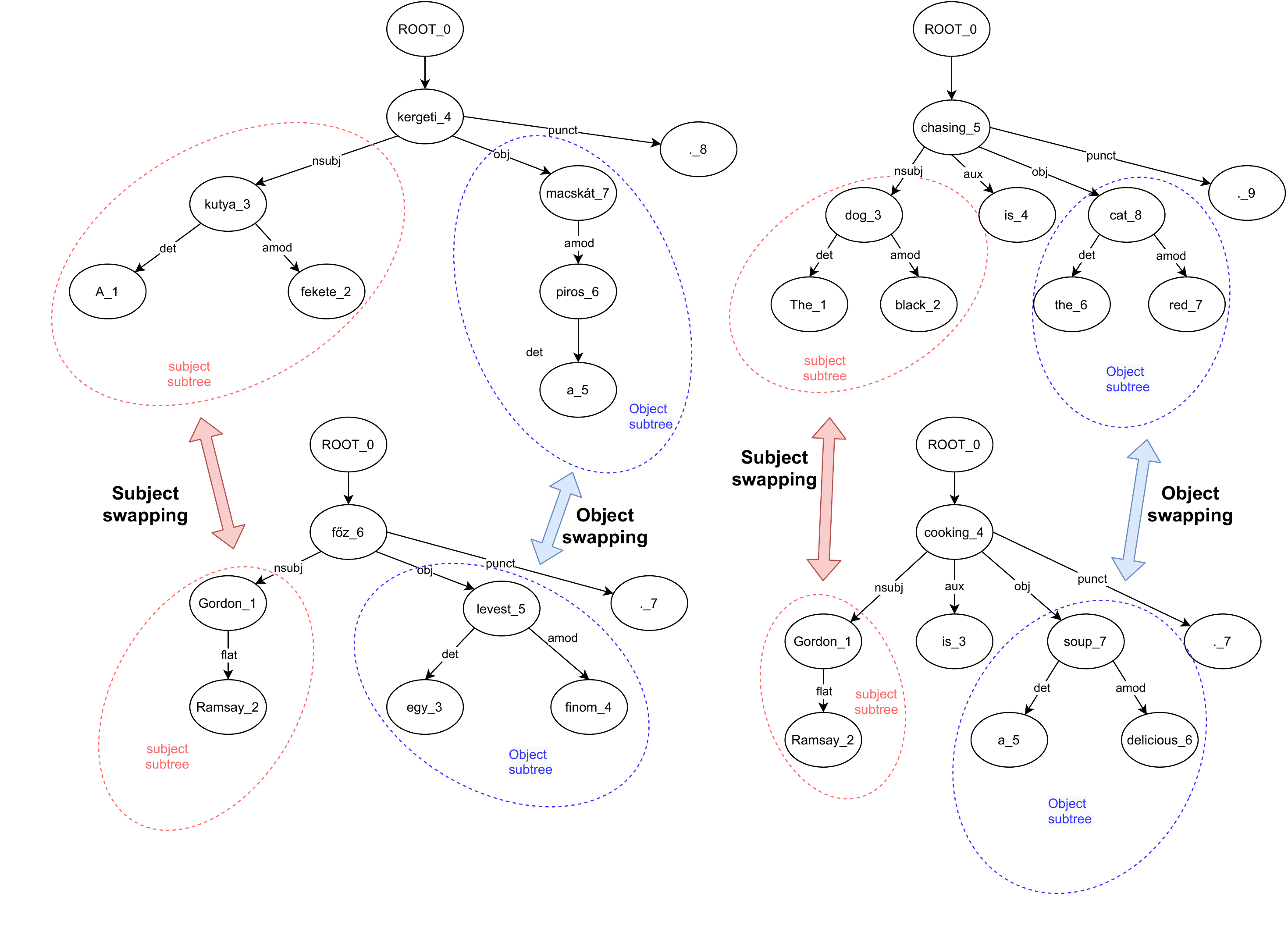}
    \caption{The process of swapping subject and object dependency subtrees between bisentences. The augmented sentences can be seen in Table \ref{table:obj-aug-example}.}
    \label{fig:syntax-aug-diagram}
}
}
\end{figure}

Finding any subtree-pair, which correspond to the same part of the source and target sentence is a very hard task, so we limit our work to finding two substructures that are common across languages: subjects and objects (see Figure \ref{fig:syntax-aug-diagram} and Table \ref{table:obj-aug-example}). As the problem space of new sentence-pairs that we can generate explodes quickly with respect to the size of the dataset, we filter sentences based on their dependency parse trees with two conditions. First, the dependency trees must contain only one subject and object for both the source and target sentences. Second, the subtrees corresponding to objects and subjects must be a consecutive sequence of words with respect to the original word order. In our experiments, we found that about 5\% of the sentence pairs satisfy the above two conditions and therefore are eligible for augmentation. It is convenient to have the same dependency relations for both English and Hungarian, so we use the Universal Dependencies\footnote{https://universaldependencies.org/} \citep{nivre2016universal} tag set. Implementation-wise, we chose the Stanza\footnote{https://stanfordnlp.github.io/stanza/} dependency parser \citep{qi2020stanza} for English and the parser in the Hungarian Spacy\footnote{https://github.com/huspacy/huspacy} model for Hungarian.

We perform a series of simulated low-resource experiments on a subcorpus of Hunglish2. We constrain our experiment to only one subcorpus, because this better simulates the lack of diversity in parallel corpora of low-resource languages. We subsample from the Modern Literature subcorpus of Hunglish2 from 5k to 500k. For each dataset, we perform three experiments: train a baseline Transformer model and train the same model with additional data from the two proposed augmentation methods. In every iteration, we extend the datasets with 50\% augmented data. Every model is evaluated on the same held-out test set, that is a fraction of our original test set with samples only from the Modern Literature subcorpus.

We acknowledge that sampling from a medium- or high-resource corpus is not the same as working on a truly low-resource language pair. The latter is likely a much worse representation of the overall population, but for the course of this work we limit our experiments to Hungarian-English. Extending these experiments to actual low-resource language pairs is a promising direction for future work.

\begin{table}[!htbp]
\centering
\begin{tabular}{l|l}
Sentence1-EN & The black dog is chasing the red cat.      \\
Sentence1-HU & A fekete kutya kergeti a piros macskát.    \\
Sentence2-EN & Gordon Ramsay is cooking a delicious soup. \\
Sentence2-HU & Gordon Ramsay egy finom levest főz.        \\ \hline
EN-OBJ-AUG-1     & The black dog is chasing a delicious soup. \\
HU-OBJ-AUG-1     & A fekete kutya kergeti egy finom levest.    \\
EN-OBJ-AUG-2     & Gordon Ramsay is cooking the red cat.      \\
HU-OBJ-AUG-2     & Gordon Ramsay a piros macskát főz.        
\end{tabular}
\caption{Augmentation via subtree swapping of objects.}
\label{table:obj-aug-example}
\end{table}

\section{Results}
In this section, we provide a detailed evaluation of both the Transformer-based machine translation models and the proposed augmentation method. For quantitative evaluation, we use the BLEU score.

\subsubsection{NMT models for EN-HU and HU-EN} Our baseline model without augmentation achieved a BLEU score of \textbf{33.4} for EN-HU and \textbf{40.0} for HU-EN on the held-out test set. In Table \ref{table:example-translations-best-models}, we collected a few example translations from the test set provided by our best models.


\begin{table}[!htbp]
\centering
\begin{tabularx}{\textwidth}{c>{\raggedright}X}
\toprule
\textbf{Example\ \#} & \textbf{Sentence} \tabularnewline
\midrule
1 & \textbf{Source} \\
    Villefort ezeket az utolsó szavakat olyan lázas dühvel ejtette ki, ami egészen vadul ékesen szólóvá tette. \\
    \textbf{Reference} \\
    Villefort pronounced these last words with a feverish rage, which gave a ferocious eloquence to his words. \\
    \textbf{Predicted} \\
    Villefort pronounced these last words with a feverish rage which rendered him passionately eloquent. \tabularnewline

\midrule
2 & \textbf{Source} \\
    Amióta Merytonba szállásolták az ezredet, csak a szerelem, az udvarlás, a tisztek jártak az eszében. \\
    \textbf{Reference} \\
    Since the shire were first quartered in Meryton, nothing but love, flirtation, and officers have been in her head. \\
    \textbf{Predicted} \\
    He had been thinking of love, of courting, of officers, ever since the regiment came to Meryton. \tabularnewline
\midrule
3 & \textbf{Source} \\
    Malfoy, Crak és Monstro Csikócsőrrel próbálkoztak.\\
    \textbf{Reference} \\
    Malfoy, Crabbe, and Goyle had taken over Buckbeak. \\
    \textbf{Predicted} \\
    Malfoy, Crabbe, and Goyle had tried Buckbeak. \tabularnewline
    \midrule \midrule
4 & \textbf{Source} \\
    His remembrance shall be sweet as honey in every mouth, and as music at a banquet of wine. \\
    \textbf{Reference} \\
    Mint ínyünknek a méz, édes az emléke, vagy mint a nótaszó borozgatás közben. \\
    \textbf{Predicted} \\
    Emlékezete édes lesz, mint a méz minden szájban, és mint a zene a bor lakomáján. \tabularnewline

\midrule
5 & \textbf{Source} \\
    His eyes moved toward the hunting knife that had been slung over the mosquito-net bar by the dead man the day he arrived. \\
    \textbf{Reference} \\
    Szeme a vadászkésre siklott, amelyet a halott ember a moszkitóháló keretére dobott azon a napon, amikor megérkezett. \\
    \textbf{Predicted} \\
    Szeme a vadászkés felé fordult, melyet a halott férfi a moszkitóháló rácsára dobott az érkezése napján. \tabularnewline
\midrule
6 & \textbf{Source} \\
    He was frozen stiff in the weeds beside the track. \\
    \textbf{Reference} \\
    Csonttá fagyva feküdt a vágány mellett a gazos földön. \\
    \textbf{Predicted} \\
    Merevre fagyott a vágány melletti gazban. \tabularnewline
\bottomrule
\end{tabularx}
\caption{Example translations produced by the best EN-HU and HU-EN translator models.}
\label{table:example-translations-best-models}
\end{table}

\subsubsection{Syntax-aware data augmentation}
\begin{figure}[!htbp]
\centerline{
\scalebox{0.9}{
    \includegraphics[width=0.88\textwidth]{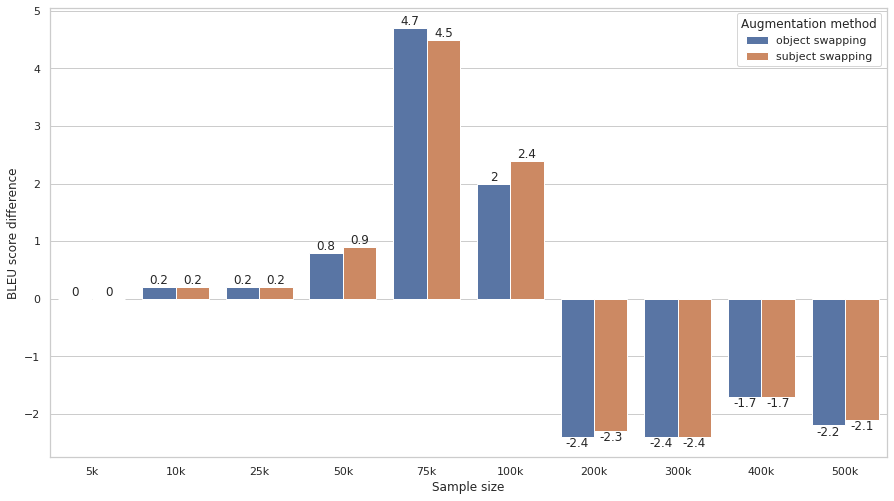}
    \caption{Absolute BLEU score differences compared to the baseline model for each sample size. (EN-HU)}
    \label{fig:low-res-bleu-score-diffs-enhu}
}
}
\end{figure}

\begin{figure}[!htbp]
\centerline{
\scalebox{0.9}{
    \includegraphics[width=0.88\textwidth]{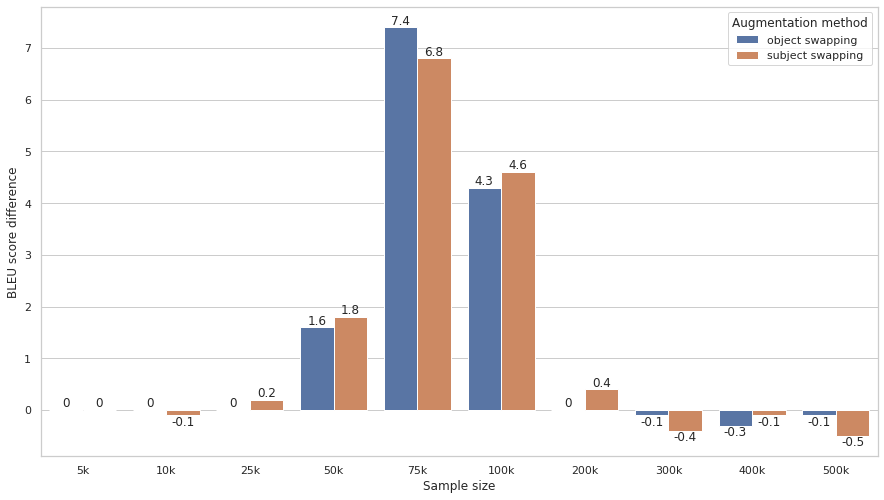}
    \caption{Absolute BLEU score differences compared to the baseline model for each sample size. (HU-EN)}
    \label{fig:low-res-bleu-score-diffs-huen}
}
}
\end{figure}

The scores of our simulated low-resource experiments can be found in Table \ref{table:low-resource-results} for both HU-EN and EN-HU. The absolute performance gain due to the augmentation is visualized in Figures \ref{fig:low-res-bleu-score-diffs-enhu} and \ref{fig:low-res-bleu-score-diffs-huen} for each sample size. With a smaller sample size (5k, 10k, 25k), the models had a close to 0 BLEU score. This can likely be contributed to the fact that the model is fixed for all experiments and it is probably too complex for datasets of this size regardless of augmentation. In the 50k-100k range, we observe visible improvements in the BLEU score of models trained with augmentation compared to the baseline models. With a 75k sample size, the baseline BLEU scores of 0.9 and 1.4 are significantly outperformed by the model trained with object swapping augmentation with BLEU scores of 6.1 and 8.8 for EN-HU and HU-EN respectively. In the 100k-500k range, we do not see improvement in BLEU score with the augmentation methods. With two exceptions, the models using augmented data perform slightly worse than the baseline model. As the augmentation ratio is also fixed to 0.5 during all experiments, it is possible that at this scale we inject too many and too noisy new data points into the training set. We also examined the reason behind the noisiness of the augmentation by manual analysis. We collected a few common error types, which are listed in Table \ref{table:augmentation-common-error-types}. Apart from the ones listed below, we found that most of the errors propagate from an incorrect dependency parsing, especially for Hungarian. We observed many falsely identified subjects, especially in cases, where the subject of the sentence was dropped (pronoun-dropping).

\begin{table}[!htbp]
\centering
\begin{tabularx}{\textwidth}{c>{\raggedright}X}
\toprule
\textbf{Error type\ } & \textbf{Sentences} \tabularnewline
\midrule
Article definiteness & \textbf{Source (aug)} \\
    Outside the apothecary, Hagrid checked \textcolor{red}{the} weapons again. \\
    \textbf{Target (aug)} \\
    Mikor végeztek a patikában, Hagrid még egyszer ellenőrizte \textcolor{red}{a} fegyvert. \tabularnewline
\midrule
Coreference & \textbf{Source (aug)} \\
    Two other companies claimed only \textcolor{red}{it}. \\
    \textbf{Target (aug)} \\
    Két másik vállalat csak \textcolor{red}{egyéni elbánást} kérelmezett. \tabularnewline
\midrule
Conjugation & \textbf{Source (aug)} \\
    Member states had also played an ironic role here. \\
    \textbf{Target (aug)} \\
    A tagállamok ismét ironikus játékot \textcolor{red}{űzött} vele. \tabularnewline
\midrule
Different subject & \textbf{Source (aug)} \\
    \textcolor{red}{Captain}, how many men did the wind leave on mars? \\
    \textbf{Target (aug)} \\
    \textcolor{red}{A szél}, hány embert hagytál a Marson? \tabularnewline
\midrule
Pronoun dropping & \textbf{Source (orig)} \\
    It shall submit the reports to the european parliament and to the council. \\
    \textbf{Target (orig)} \\
    \textcolor{red}{Ő} A jelentéseket az Európai Parlamenthez és a Tanácshoz nyújtja be. \tabularnewline
\bottomrule
\end{tabularx}
\caption{Common error types of the syntax-aware data augmentation.}
\label{table:augmentation-common-error-types}
\end{table}

\begin{table}[!htbp]
\centering
\begin{tabular}{|c|l|c|c|}
\hline
\textbf{Sample size} & \multicolumn{1}{l|}{\textbf{Method}} & \multicolumn{1}{l|}{\textbf{EN-HU BLEU}} & \multicolumn{1}{l|}{\textbf{HU-EN BLEU}} \\ \hline
\multirow{3}{*}{5k} & base & 0.1 & 0.0 \\ \cline{2-4}
 & object swapping & 0.0 & 0.0 \\ \cline{2-4}
 & subject swapping & 0.0 & 0.0 \\ \hline \hline
\multirow{3}{*}{10k} & base & 0.2 & 0.1 \\ \cline{2-4}
& object swapping & \textbf{0.3} & 0.1 \\ \cline{2-4}
& subject swapping & \textbf{0.3} & 0.0 \\ \hline \hline
\multirow{3}{*}{25k} & base & 0.4 & 0.1 \\ \cline{2-4}
& object swapping & 0.3 & 0.1 \\ \cline{2-4}
& subject swapping & 0.3 & \textbf{0.3} \\ \hline \hline
\multirow{3}{*}{50k} & base & 0.4 & 0.8 \\ \cline{2-4}
& object swapping & \textbf{1.6} & \textbf{2.4} \\ \cline{2-4}
& subject swapping & \textbf{1.7} & \textbf{2.6} \\ \hline \hline
\multirow{3}{*}{75k} & base & 0.9 & 1.4 \\ \cline{2-4}
& object swapping & \textbf{6.1} & \textbf{8.8} \\ \cline{2-4}
& subject swapping & \textbf{5.9} & \textbf{8.2} \\ \hline \hline
\multirow{3}{*}{100k} & base & 3.0 & 5.4 \\ \cline{2-4}
& object swapping & \textbf{7.4} & \textbf{9.7} \\ \cline{2-4}
& subject swapping & \textbf{7.8} & \textbf{10.0} \\ \hline \hline
\multirow{3}{*}{200k} & base & 12.7 & 14.5 \\ \cline{2-4}
& object swapping & 12.1 & 14.5 \\ \cline{2-4}
& subject swapping & 12.2 & \textbf{14.9} \\ \hline \hline
\multirow{3}{*}{300k} & base & 14.4 & 16.5 \\ \cline{2-4}
& object swapping & 14.1 & 16.4 \\ \cline{2-4}
& subject swapping & 14.1 & 16.1 \\ \hline \hline
\multirow{3}{*}{400k} & base & 15.5 & 17.0 \\ \cline{2-4}
& object swapping & 15.3 & 16.7 \\ \cline{2-4}
& subject swapping & 15.3 & 16.9 \\ \hline \hline
\multirow{3}{*}{500k} & base & 15.5 & 17.7 \\ \cline{2-4}
& object swapping & 15.5 & 17.6 \\ \cline{2-4}
& subject swapping & \textbf{15.6} & 17.2 \\ \hline 
\end{tabular}
\caption{Results of our low-resource experiments. The BLEU scores in bold indicate an experiment, where the model with augmentation outperformed the baseline for that particular sample.}
\label{table:low-resource-results}
\end{table}

\section{Conclusion}
We presented Transformer-based NMT models for Hungarian-English and English-Hungarian. Our best models achieve a BLEU score of \textbf{40.0} and \textbf{33.4} for HU-EN and EN-HU respectively. We also shared results of an ongoing work on a potential data augmentation method alternative to back-translation in lower-resource scenarios. We briefly discussed this syntax-aware method, which creates new data points by swapping specific subtrees of dependency parse trees in parallel for both the source and target sentences. Regarding our future work, we plan to fix some of the common errors listed in Table \ref{table:augmentation-common-error-types} and therefore enhance the augmentation technique by making the generated samples less noisy. We also plan to extend our experiments to other languages.


\newpage

\section*{Acknowledgements}
\label{Koszonet}
The authors would like to thank András Kornai for discussions on the syntax-aware data augmentation.

%
\renewcommand\bibname{References}
\bibliographystyle{splncsnat_en}
\bibliography{mszny}

\end{document}